\documentclass[10pt,twocolumn,letterpaper]{article}

\usepackage[pagenumbers]{cvpr} 

\usepackage[accsupp]{axessibility}
\usepackage{graphicx}
\usepackage{float}
\usepackage{amsmath}
\usepackage{amssymb}
\usepackage{booktabs}
\usepackage[table]{xcolor}
\usepackage{multirow, overpic, textpos}

\usepackage[pagebackref,breaklinks,colorlinks]{hyperref}

\usepackage[capitalize]{cleveref}
\crefname{section}{Sec.}{Secs.}
\Crefname{section}{Section}{Sections}
\Crefname{table}{Table}{Tables}
\crefname{table}{Tab.}{Tabs.}

\def\cvprPaperID{278} 
\def\confName{CVPR}
\def\confYear{2023}

\newlength\savewidth\newcommand\shline{\noalign{\global\savewidth\arrayrulewidth\global\arrayrulewidth 1pt}\hline\noalign{\global\arrayrulewidth\savewidth}}
\newcommand{\tablestyle}[2]{\setlength{\tabcolsep}{#1}\renewcommand{\arraystretch}{#2}\centering\footnotesize}
\renewcommand{\paragraph}[1]{\vspace{1.25mm}\noindent\textbf{#1}}

\newcolumntype{x}[1]{>{\centering\arraybackslash}p{#1pt}}
\newcolumntype{y}[1]{>{\raggedright\arraybackslash}p{#1pt}}
\newcolumntype{z}[1]{>{\raggedleft\arraybackslash}p{#1pt}}

\begin{document}

\title{Masked Image Modeling with Local Multi-Scale Reconstruction}

\author{
	Haoqing Wang\textsuperscript{\rm 1},
	Yehui Tang\textsuperscript{\rm 1,2},
        Yunhe Wang\textsuperscript{\rm 2}\footnotemark[1],
	Jianyuan Guo\textsuperscript{\rm 2},
	Zhi-Hong Deng\textsuperscript{\rm 1}\footnotemark[1],
	  Kai Han\textsuperscript{\rm 2} \\
	\textsuperscript{\rm 1}National Key Lab of General AI, School of Intelligence Science and Technology, Peking University \\
	\textsuperscript{\rm 2}Huawei Noah’s Ark Lab \\
	\{wanghaoqing,yhtang,zhdeng\}@pku.edu.cn, \{kai.han, yunhe.wang\}@huawei.com
}

\maketitle

\begin{abstract}
Masked Image Modeling (MIM) achieves outstanding success in self-supervised representation learning. Unfortunately, MIM models typically have huge computational burden and slow learning process, which is an inevitable obstacle for their industrial applications. Although the lower layers play the key role in MIM, existing MIM models conduct reconstruction task only at the top layer of encoder. The lower layers are not explicitly guided and the interaction among their patches is only used for calculating new activations. Considering the reconstruction task requires non-trivial inter-patch interactions to reason target signals, we apply it to multiple local layers including lower and upper layers. Further, since the multiple layers expect to learn the information of different scales, we design local multi-scale reconstruction, where the lower and upper layers reconstruct fine-scale and coarse-scale supervision signals respectively. This design not only accelerates the representation learning process by explicitly guiding multiple layers, but also facilitates multi-scale semantical understanding to the input. Extensive experiments show that with significantly less pre-training burden, our model achieves comparable or better performance on classification, detection and segmentation tasks than existing MIM models. Code is available with both \href{https://gitee.com/mindspore/hub/blob/fa2a3270aa36673f835e524fa55c5a4c67262eb2/mshub_res/assets/noah-cvlab/gpu/1.8/localmim_v1.0_imagenet2012.md}{MindSpore} and \href{https://github.com/huawei-noah/Efficient-Computing/tree/master/Self-supervised/LocalMIM}{PyTorch}.

\end{abstract}

\renewcommand{\thefootnote}{\fnsymbol{footnote}}
\footnotetext[1]{Corresponding author.}

\section{Introduction}
Recently, Masked Image Modeling (MIM) \cite{bao2021beit,he2022masked,xie2022simmim} achieves outstanding success in the field of self-supervised visual representation learning, which is inspired by the Masked Language Modeling (MLM) \cite{kenton2019bert,brown2020language} in natural language processing and benefits from the development of vision transformers \cite{DBLP:conf/iclr/DosovitskiyB0WZ21,liu2021swin,wang2021pyramid}. MIM learns semantic representations by first masking some parts of the input and then predicting their signals based on the unmasked parts, e.g., normalized pixels \cite{he2022masked,xie2022simmim}, discrete tokens \cite{bao2021beit,dong2021peco}, HOG feature \cite{wei2022masked}, deep features \cite{baevski2022data2vec,zhou2021image} or frequencies \cite{xie2022masked,liu2022devil}.

Despite superior performance on various downstream tasks, these models have huge computational burden and slow learning process \cite{huang2022green}. They typically require thousands of GPU Hours for pre-training on ImageNet-1K to get generalizing representations. Since we expect to pre-train these models on more massive amount of unlabeled data (e.g., free Internet data) to obtain more generalizing representations in practice, the pre-training efficiency is an inevitable bottleneck limiting the industrial applications of MIM. How to accelerate the representation learning in MIM is an important topic. To this end, MAE \cite{he2022masked} pioneered the asymmetric encoder-decoder strategy, where the costly encoder only operates few visible patches and the lightweight decoder takes all the patches as input for prediction. Further, GreenMIM \cite{huang2022green} extends the asymmetric encoder-decoder strategy to hierarchical vision transformers (e.g., Swin \cite{liu2021swin}). Besides, \cite{chen2022efficient,li2022uniform,guo2022fastmim} shrinks the input resolution to lessen the input patches, thereby reducing the computational burden. However, they all aim to accelerate the encoding process rather than the representation learning.

\begin{figure*}
	\centering
	\subcaptionbox{ViT-B}{
		\includegraphics[width=0.45\textwidth]{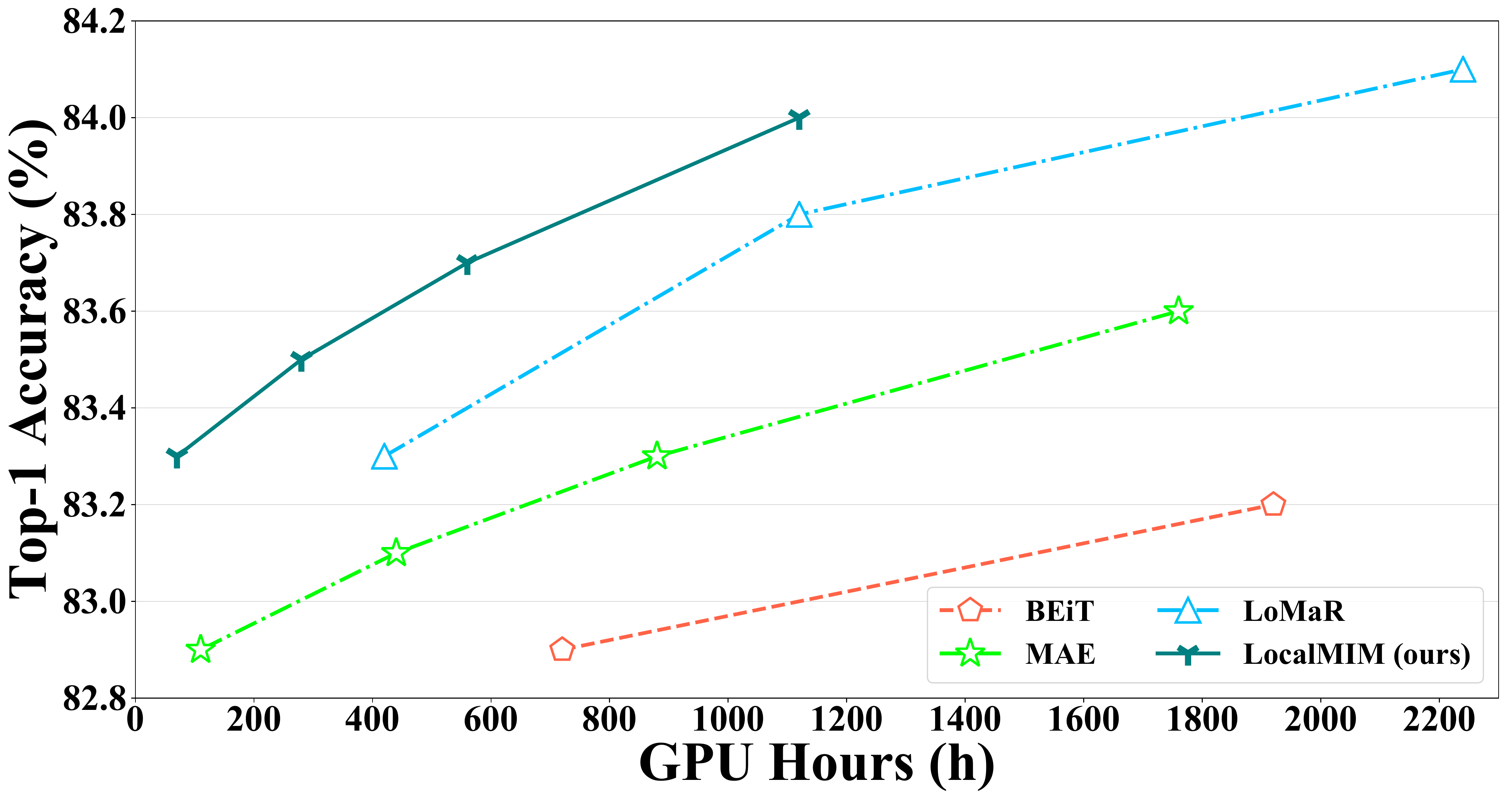}}
	\subcaptionbox{Swin-B}{
		\includegraphics[width=0.45\textwidth]{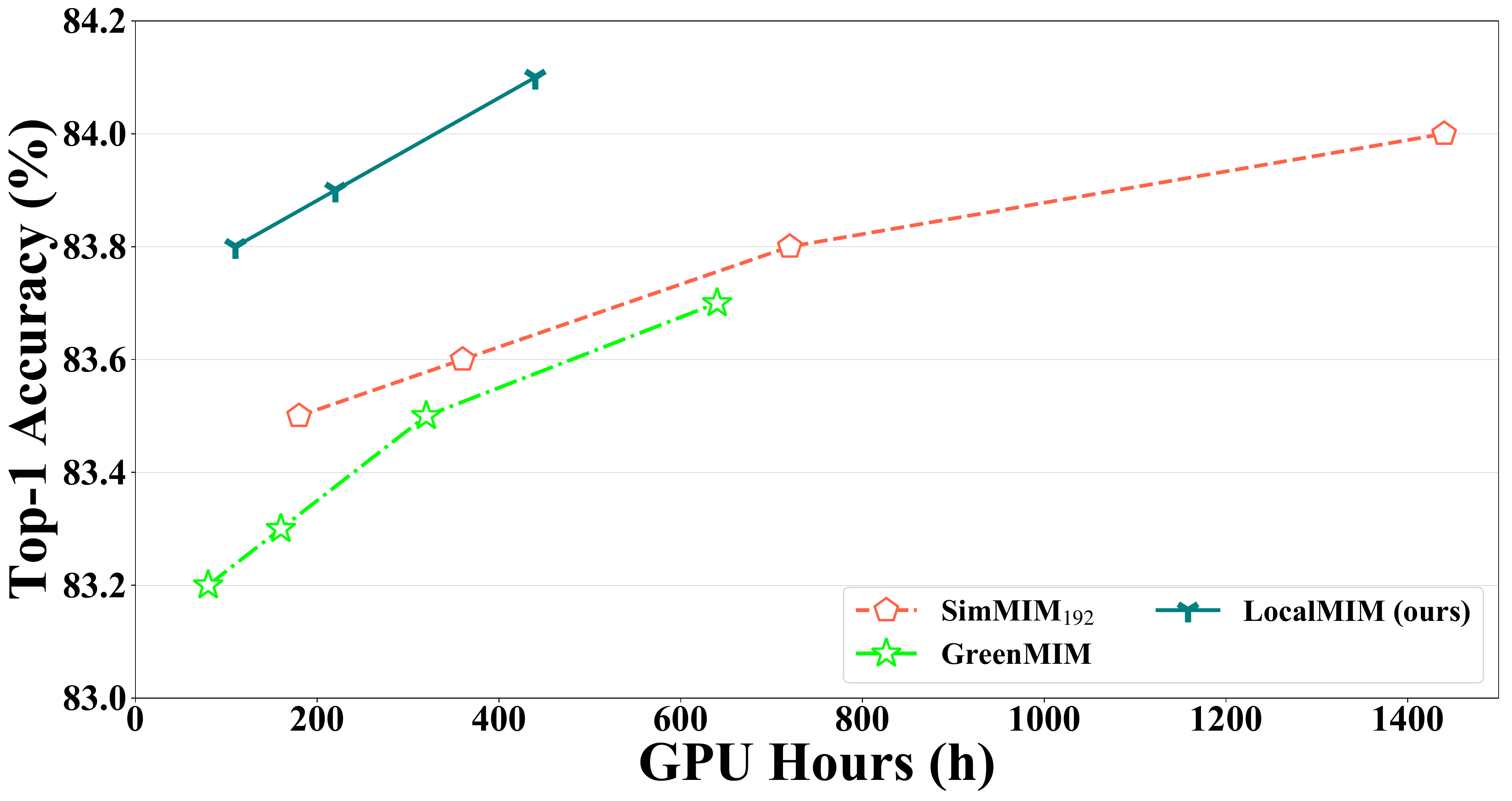}}
	\caption{Top-1 fine-tuning accuracy on ImageNet-1K vs. Pre-training duration. The duration is estimated on a machine with one Tesla V100-32G GPU, CUDA 10.2 and PyTorch 1.8. `GPU Hours' is the running time on single GPU.}
	\label{trade}
	\vspace{-3mm}
\end{figure*}

In MIM, the learning of upper layers depends on that of lower ones during pre-training, since the upper-layer features are calculated from the lower layers. Besides, during fine-tuning the upper layers are typically tuned quickly to adapt to the downstream task while the lower ones change more slowly and need to be well-learned \cite{zhang2020revisiting,howard2018universal,bao2021beit}. Even fine-tuning only the several upper layers and freezing the others can obtain similar performance \cite{he2022masked}. Therefore, the lower layers of encoder play the key role in MIM. However, all existing MIM models only conduct reconstruction task at the top layer of encoder and the lower ones are not explicitly guide, thus the interaction among their patches is only used for calculating the activations of the next layer. Considering the reconstruction task requires non-trivial inter-patch interactions to reason target signals, we apply it to both lower and upper layers to explicitly guide them and thus accelerate the overall learning process. Using tiny decoder is sufficient for each local reconstruction task and does not significantly increase the computational burden.

\begin{figure*}
	\centering
	\subcaptionbox{Overview of LocalMIM.}{
		\includegraphics[width=0.675\textwidth]{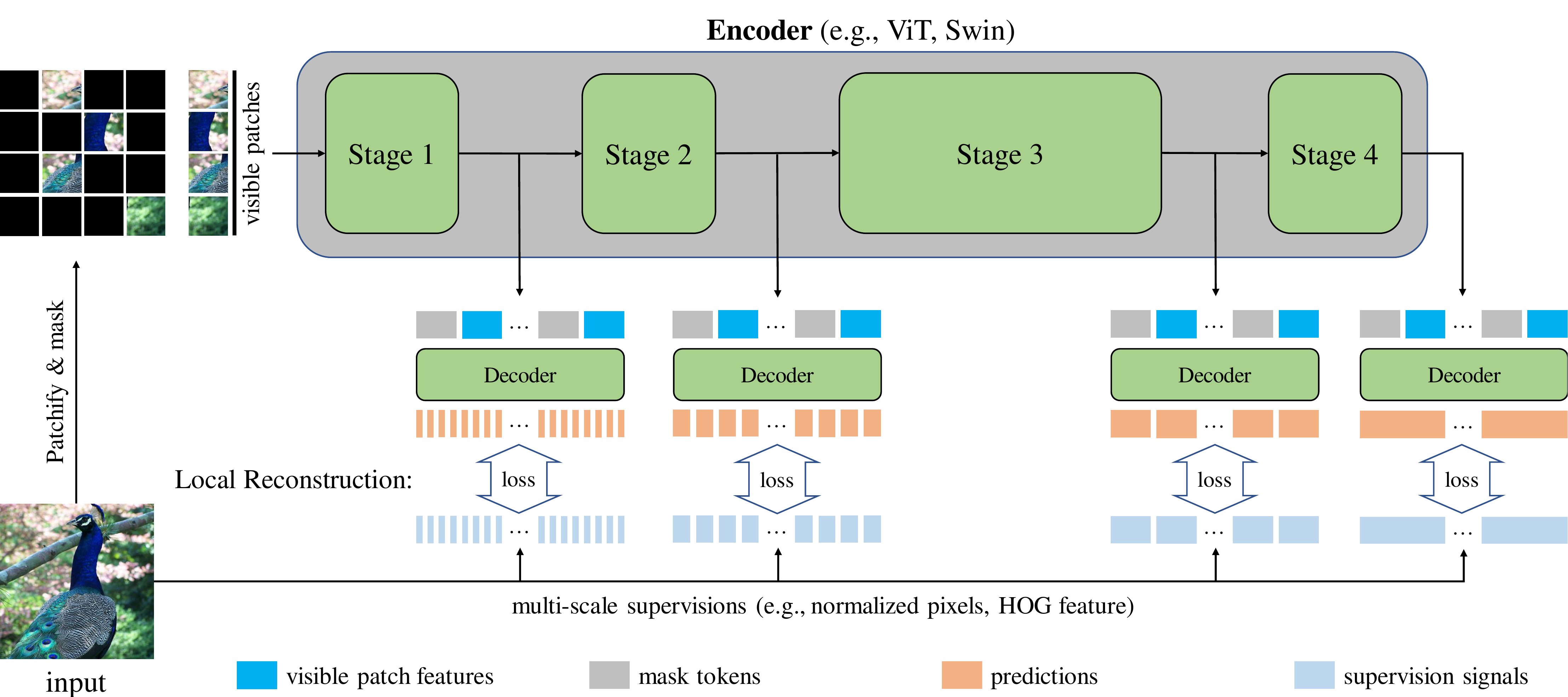}}
	\subcaptionbox{Decoding process under a scale.}{
		\includegraphics[width=0.3\textwidth]{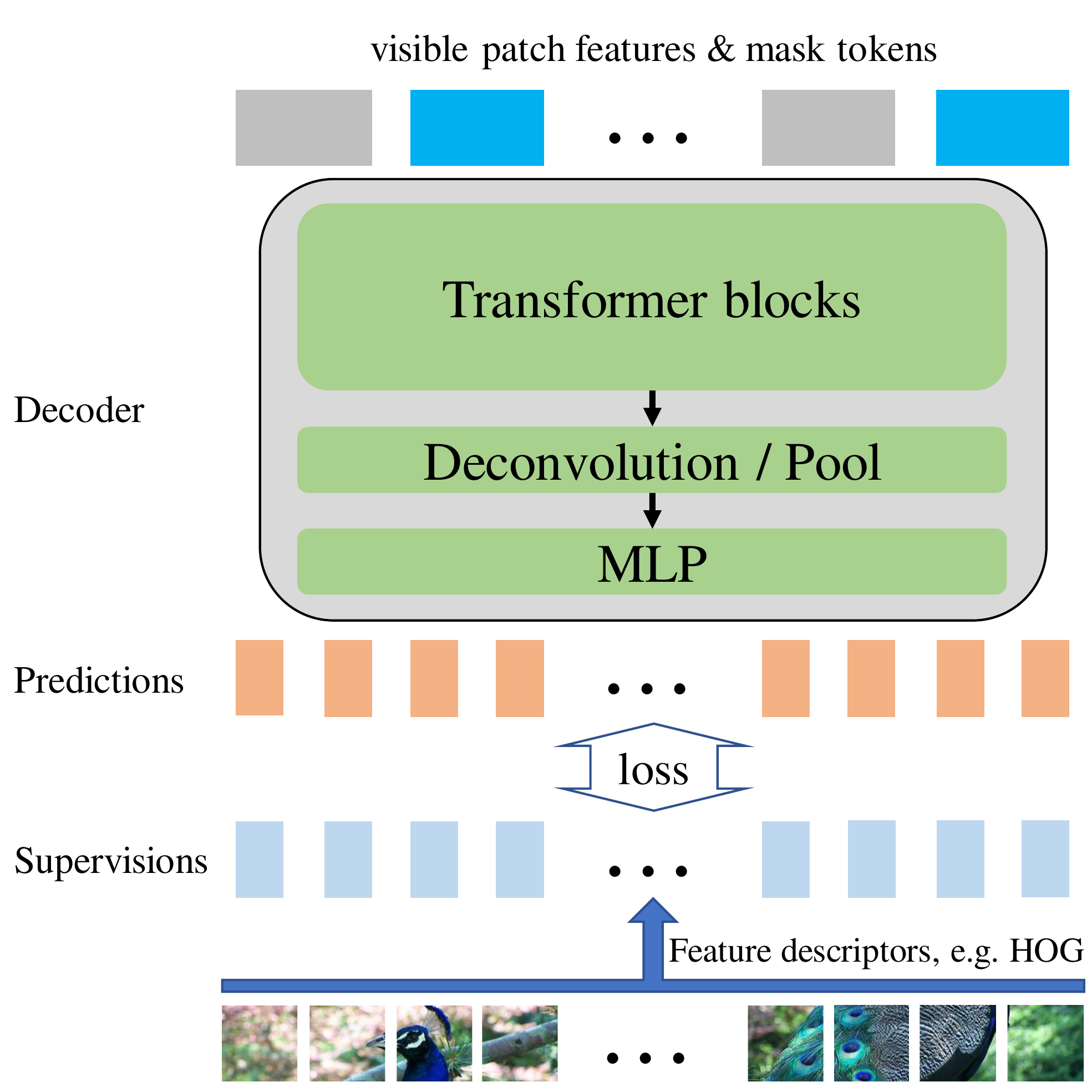}}
	\vspace{-2mm}
	\caption{Illustration of local multi-scale reconstruction. a) We randomly mask some input patches and then predict their supervision signals of different scales at multiple local layers. The multi-scale supervisions are obtained by first dividing the input under different scales and then extracting signals with some feature descriptors. The lower layers are responsible for fine-scale reconstruction and the upper ones are responsible for coarse-scale reconstruction. We also use the asymmetric encoder-decoder strategy for quick encoding. b) The decoder consists of three parts: Transformer blocks, Deconvolution/Pool (optional) and Multilayer perceptron.}
	\label{model}
	\vspace{-3mm}
\end{figure*}

How to properly conduct reconstruction tasks at multiple local layers is a non-trivial problem. For example, applying the top-layer reconstruction task to carefully chosen local layers of ViT \cite{DBLP:conf/iclr/DosovitskiyB0WZ21} can not achieve meaningful improvement. In general, the lower layers exploit low-level information and the upper ones learn high-level information \cite{park2021vision,gao2022convmae}, so it is not appropriate to use the supervision signals of same scale for multiple local reconstruction tasks. Here 'scale' is the spatial size of the supervision signals calculated from the divided input regions, e.g., the signals from the $p\times p$ regions in an input of $H\times W$ resolution has the scale of $\frac{H}{p}\times\frac{W}{p}$. The fine-scale and coarse-scale supervisions typically contain low-level and high-level information of the input respectively, and these multi-scale supervisions from input are widely ignored by existing MIM models. To this end, we propose local multi-scale reconstruction where the lower and upper layers reconstruct fine-scale and coarse-scale supervisions respectively. This design not only accelerates the representation learning process, but also facilitates multi-scale semantical understanding to the input. When the decoded predictions have different scale with the supervisions (e.g., on ViT), we use the deconvolution/pool operations to rescale them. We also apply the asymmetric encoder-decoder strategy \cite{he2022masked,huang2022green} for quick encoding. Our model, dubbed as LocalMIM, are illustrated in Fig. \ref{model} (a).

Overall, we summarize our contributions as follows.
\begin{itemize}
\item To the best of our knowledge, this is the first work in MIM to conduct local reconstructions and use multi-scale supervisions from the input.
\item Our model is architecture-agnostic and can be used in both columnar and pyramid architectures.
\item From extensive experiments, we find that 1) LocalMIM is more efficient than existing MIM models, as shown in Fig. \ref{trade} and Table \ref{Finetune}. For example, LocalMIM achieves the best MAE result with $3.1\times$ acceleration on ViT-B and the best GreenMIM result with $6.4\times$ acceleration on Swin-B. 2) In terms of top-1 fine-tuning accuracy on ImageNet-1K, LocalMIM achieves $84.0\%$ using ViT-B and $84.1\%$ using Swin-B with significantly less pre-training duration than existing MIM models. The obtained representations also achieve better generalization on detection and segmentation downstream tasks, as shown in Table \ref{ADE20K} and \ref{COCO}.
\end{itemize}

\section{Related Works}
\paragraph{Masked Image Modeling.}
With the development of vision transformers \cite{DBLP:conf/iclr/DosovitskiyB0WZ21,liu2021swin,wang2021pyramid,han2021transformer,sun2019fast}, Masked Image Modeling (MIM) gradually replaces the dominant position of contrastive learning \cite{he2020momentum,chen2020simple,wang2022rethinking} in visual self-supervised representation learning due to its superior fine-tuning performance in various visual downstream tasks. Many target signals have been designed for the mask-prediction pretext task in MIM, such as normalized pixels \cite{he2022masked,xie2022simmim}, discrete tokens \cite{bao2021beit,dong2021peco}, HOG feature \cite{wei2022masked}, deep features \cite{baevski2022data2vec,zhou2021image} or frequencies \cite{xie2022masked,liu2022devil}. However, they are all only applied as single-scale supervisions for reconstruction. An inevitable bottleneck for the industrial applications of MIM is that these models typically require huge computational resources and long pre-training duration. To this end, some works accelerate the encoding process via the asymmetric encoder-decoder strategy \cite{he2022masked,huang2022green} or lessening the input patches \cite{chen2022efficient,li2022uniform}. Only accelerating the encoding process sometimes doesn't really speed up the representation learning, like GreenMIM vs. SimMIM$_{192}$ in Fig. \ref{trade}. In this work, we use local multi-scale reconstructions to explicitly guide multiple lower layers and thus accelerate the overall learning process. Our method is compatible with the above quick encoding approaches. ConvMAE \cite{gao2022convmae} fuses the feature of local layers for the final reconstruction to explicitly guide them, but still applies single-scale supervision signals.

\paragraph{Locally supervised learning.}
Considering the biological brain learns mainly based on local information \cite{caporale2008spike}, some works \cite{lowe2019putting,nokland2019training,wang2020revisiting,pyeon2020sedona} used local error signals to greedily optimize individual blocks of the backbone without global back-propagation. This greedy training procedure significantly reduces memory overhead, but can not accelerate model learning. In this work, we also optimize error at multiple local layers but still use global back-propagation for end-to-end training. Compared with existing MIM models, our local reconstruction is more biologically plausible since the human brain prefers local learning rules \cite{crick1989recent,dan2004spike,bengio2015towards}. Feature distillation \cite{DBLP:journals/corr/RomeroBKCGB14,heo2019comprehensive} also explicitly guides the local layers, but needs costly pre-trained or momentum teacher. In fact, the multi-scale supervisions from original input are sufficient to guide local layers and also readily available.

\paragraph{Multi-scale property.}
Biological visual perception is hierarchical and multi-scale \cite{hubel1960receptive}. Moreover, multi-scale features are also useful for many visual tasks \cite{ren2015faster,he2017mask,chen2017deeplab}. To this end, many advanced vision transformers \cite{liu2021swin,wang2021pyramid,fan2021multiscale,zhang2021multi,ijcai2021-149} hard-code the multi-scale property to the architectures and boost the performance. However, global single-scale reconstruction is not enough for guiding multiple local layers to learn multi-scale information. In this work, beyond the multi-scale features, we further introduce multi-scale supervisions to soft-code this property. This new design is not limited to specific architectures.

\section{Model}
\subsection{Masked Image Modeling}
In MIM, the models first mask some parts of the input image and then predict their information based on the unmasked observation. For quick encoding, MAE \cite{he2022masked} and GreenMIM \cite{huang2022green} use the asymmetric encoder-decoder strategy. The major designs are described below.

\paragraph{Image presentation.}
Generally, we present an image to the sequence of visual patches as the input to vision transformers \cite{DBLP:conf/iclr/DosovitskiyB0WZ21,liu2021swin}. The input image $x\in\mathbb{R}^{H\times W\times C}$ is reshaped to $N=HW/p^2$ non-overlapping patches $x^p\in\mathbb{R}^{N\times(p^2C)}$, where $p$ is the patch size, $(H,W)$ is the resolution of input image and $C$ is the number of channels. The patches $\{x_i^p\}_{i=1}^N$ are then linearly mapped to the patch embeddings. For retaining the positional information, the patches are typically added with positional embeddings.

\paragraph{Vision transformers.}
Unlike the grid operations in the convolutional neural networks, vision transformers \cite{DBLP:conf/iclr/DosovitskiyB0WZ21,liu2021swin,wang2021pyramid} use the stacked multi-head self-attention modules. To obtain the multi-scale feature maps for dense prediction tasks, some advanced vision transformers \cite{liu2021swin,wang2021pyramid} extend the columnar ViT \cite{DBLP:conf/iclr/DosovitskiyB0WZ21} to the pyramidal structure, where the feature maps change from fine-scale to coarse-scale and are expected to capture the multi-scale information from the input. The patches $[x_1^p,\cdots,x_N^p]$ are fed to a vision transformer with $L$ layers and obtain the output feature at $l$-th layer $z^l=[z_1^l,\cdots,z_{N_l}^l],l=1,\cdots,L$. 

\paragraph{Masking.}
Given the patch sequence $\{x_i^p\}_{i=1}^N$, MIM constructs a random mask $m\in\{0,1\}^N$ to indicate the masked patches that correspond to $m_i=1$. There are two main masking strategies, \emph{random masking} \cite{he2022masked} and \emph{block-wise masking} \cite{bao2021beit}. Specially, since the number of patches $N_l$ decreases through layers in the pyramidal architectures, some works \cite{huang2022green,gao2022convmae} first construct the random mask $m^L$ with the size of $N_L$, and then up-sample it to size $N$.

\paragraph{Encoder.}
Only the visible patches $x^v=\{x_i^p|m_i=0\}$ are fed to the encoder and mapped to the latent features $z_v^l,l=1,\cdots,L$, which significantly reduces compute and memory. Specially, since the local window attention in the pyramidal architectures is not compatible with the incomplete input patches (i.e., only visible patches), GreenMIM \cite{huang2022green} proposes the Optimal Grouping algorithm and the Group Window Attention scheme. After pre-training, the encoder is used in various downstream tasks.

\paragraph{Decoder.}
The decoder takes as the input of both encoded visible patches $z_v^L$ and mask tokens $\{e_{[\mathbb{M}]}|m_i^L=1\}$, where the mask token $e_{[\mathbb{M}]}$ is a shared and learnable vector. The positional embeddings are also added to them for providing the location information. Since the decoder is only used during pre-training to output the prediction $\hat{y}^L$, it can be any architectures which support the global information propagation among patches, e.g., a series of ViT or Swin blocks. Many works \cite{he2022masked,huang2022green,li2022uniform} show that using lightweight decoders is sufficient to learn generalizing representations.

\paragraph{Global reconstruction.}
All existing MIM models predict the supervision signal $y$ based on the final output feature $z_v^L$ of the encoder and minimize a global reconstruction loss
\begin{equation}
	\mathcal{L}_{MIM}=-\sum_{i=1}^{N_L}m_i^L\cdot \ln P(y_i|\hat{y}_i^L)
\end{equation}
where the loss is calculated on the masked patches with $m_i^L=1$, and $P(\cdot|\cdot)$ can be the Gaussian or Dirichlet distributions for regression or classification losses respectively.

\begin{table*}
	\centering
	\begin{tabular}{lcccccc}
		\toprule
		\textbf{Model} & \textbf{Backbone} & \textbf{$\#$ Params} & \textbf{PT Epoch}  & \textbf{GPU Hours/Ep.} & \textbf{Total GPU Hours} & \textbf{Acc} \\ 
		\midrule
		Scratch, ViT  & ViT-B  & 86M & 0 & 1.5 & - & 82.3 \\
		Scratch, Swin & Swin-B & 88M & 0 & 2.4 & - & 83.5 \\
		\hline
		MoCo v3 \cite{chen2021empirical} & ViT-B & 86M  & 600  & -   & -    & 83.2 \\
		DINO \cite{caron2021emerging} & ViT-B & 86M  & 300  & -   & -    & 82.8 \\
		BEiT \cite{bao2021beit}   & ViT-B & 86M & 800 & 2.4 & 1920 & 83.2 \\
		iBOT \cite{zhou2021image} & ViT-B & 86M & 400 & 10.1 & 4040 & 83.8 \\
		MAE \cite{he2022masked} & ViT-B & 86M  & 800  & 1.1 & 880 & 83.3 \\
		MAE \cite{he2022masked} & ViT-B & 86M  & 1600 & 1.1 & 1760 & 83.6 \\
		MAE \cite{he2022masked} & ViT-L & 307M & 1600 & 1.7 & 2720 & 85.9 \\
		MaskFeat \cite{wei2022masked} & ViT-B & 86M  & 1600 & 3.9 & 6240 & 84.0 \\
		CAE \cite{chen2022context} & ViT-B & 86M  & 800  & 2.8 & 2240 & 83.6 \\
		LoMaR$^{\dagger}$ \cite{chen2022efficient} & ViT-B & 86M  & 1600 & 1.4 & 2240 & 84.1 \\
		data2Vec$^{\dagger}$ \cite{baevski2022data2vec} & ViT-B & 86M  & 800  & 3.0 & 2400 & 84.2 \\
		PeCo \cite{dong2021peco} & ViT-B & 86M  & 800  & -   & -    & 84.5 \\
		\rowcolor{green!15} LocalMIM-HOG & ViT-B & 86M  & 100  & 0.7 & 70   & 83.3 \\
		\rowcolor{green!15} LocalMIM-HOG & ViT-B & 86M  & 1600 & 0.7 & 1120 & 84.0 \\
		\rowcolor{green!15} LocalMIM-HOG & ViT-L & 307M & 800  & 1.0 & 800 & 85.8 \\
		\hline
		SimMIM$_{192}$ \cite{xie2022simmim} & Swin-B & 88M  & 800 & 1.8 & 1440 & 84.0 \\
		SimMIM$_{192}$ \cite{xie2022simmim} & Swin-L & 197M & 800 & 3.0 & 2400 & 85.4 \\
		GreenMIM \cite{huang2022green} & Swin-B & 88M  & 800 & 0.8 & 640  & 83.7 \\
		GreenMIM \cite{huang2022green} & Swin-L & 197M & 800 & 1.4 & 1120 & 85.1 \\
		\rowcolor{green!15} LocalMIM-Pixel & Swin-B & 88M  & 100 & 1.0 & 100  & 83.7 \\
		\rowcolor{green!15} LocalMIM-HOG   & Swin-B & 88M  & 100 & 1.1 & 110  & 83.8 \\
		\rowcolor{green!15} LocalMIM-Pixel & Swin-B & 88M  & 400 & 1.0 & 400  & 84.0 \\
		\rowcolor{green!15} LocalMIM-HOG   & Swin-B & 88M  & 400 & 1.1 & 440  & 84.1 \\
		\rowcolor{green!15} LocalMIM-HOG   & Swin-L & 197M & 800 & 1.6 & 1280  & 85.6 \\
		\bottomrule
	\end{tabular}
	\caption{Top-1 fine-tuning accuracy on ImageNet-1K. All models are pre-trained and fine-tuned under $224\times224$ resolution except that SimMIM$_{192}$ uses $192\times192$ resolution for pre-training. ${\dagger}$ means using the relative positional encoding.}
	\label{Finetune}
	\vspace{-3mm}
\end{table*}

\subsection{Analysis and motivation}
For pre-training, the upper-layer features are calculated from the lower layers, so the well-learned lower layers can propagate semantical knowledge to the upper ones and facilitate their learning. For fine-tuning, the upper layers typically adapt quickly to specific downstream tasks, while the lower ones change more slowly and need to be sufficient learned during  pre-training \cite{zhang2020revisiting,howard2018universal,bao2021beit}. Even fine-tuning only the several upper layers and freezing the others can obtain comparable performance \cite{he2022masked}. Therefore, the lower layers of the encoder play the key role in MIM.

\begin{figure}
	\begin{center}
		\includegraphics[width=0.85\linewidth]{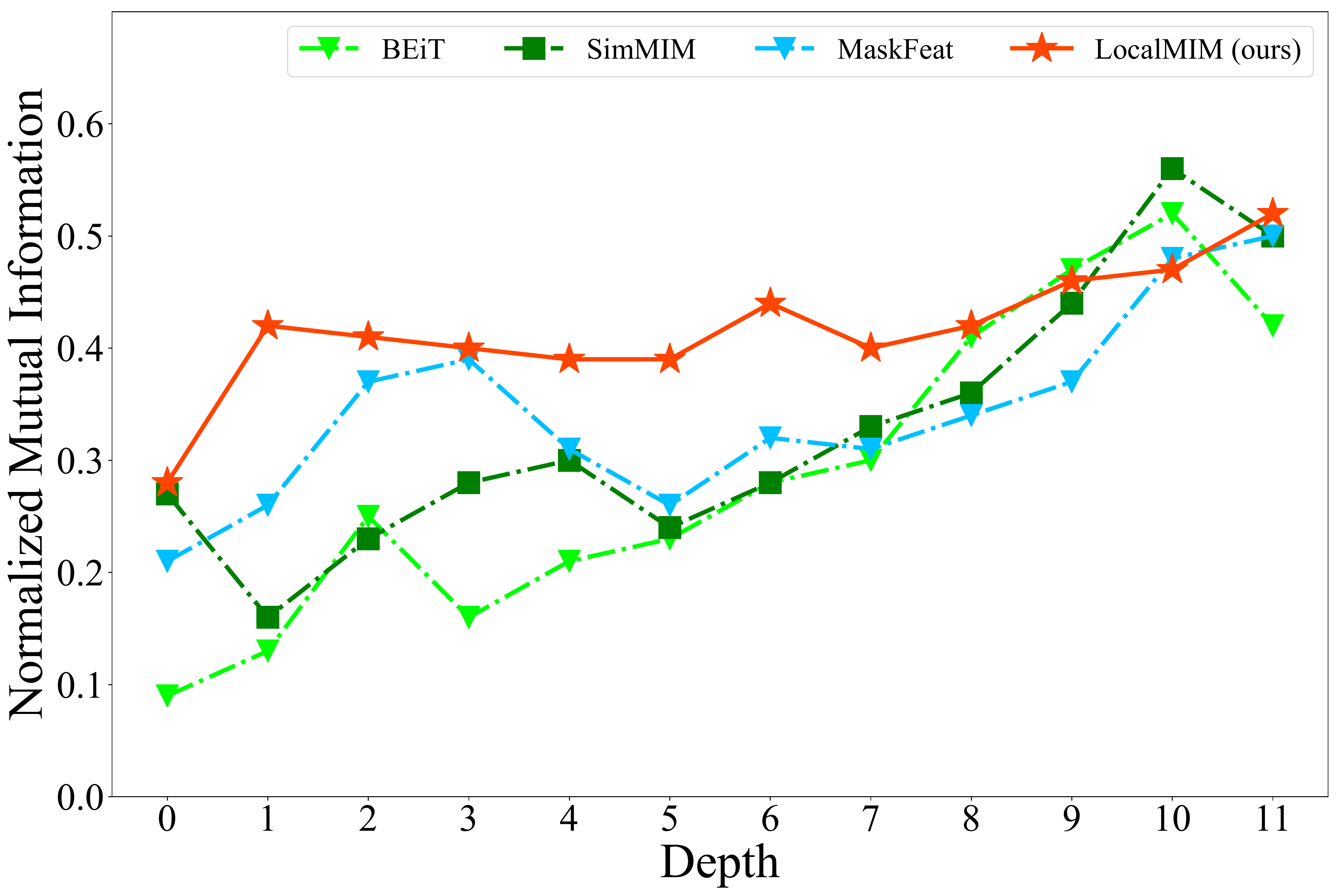}
	\end{center}
	\vspace{-3mm}
	\caption{Normalized Mutual Information (NMI) between query and key patches at each layer of a pre-trained ViT-B.}
	\label{nmi_vit}
	\vspace{-3mm}
\end{figure}

After patchification and linearly projection, the initial patch embeddings lose the inter-patch semantical relations. The self-attention mechanism in vision transformers is responsible for learning these relations by inter-patch interaction and build a better representation space than pixel space \cite{cao2022understand}. Further, since the self-attention mechanism has the computational complexity with a quadratic dependence on patch number $N$, it is difficult to learn the inter-patch interactions, especially for the lower layers of pyramidal architectures where the small patch size $p$ leads to huge $N$. However, under the global reconstruction loss, the inter-patch interaction at lower layers is not explicitly guided, and the simple task of calculating new activations is not sufficient to guide it. As a result, it is hard for the patches at lower layers to learn inter-patch relations. Concretely, we examine Normalized Mutual Information (NMI) \cite{strehl2002cluster} between query and key patches at each layer, where the high NMI value means the attention maps strongly depend on the query patches. Intuitively, the semantical representations should have highly query-adaptive attentions, i.e., the different query patches faithfully attend to their semantically related regions. This is an advantage of the self-attention mechanism. As shown in Fig. \ref{nmi_vit}, existing MIM models with global loss have small NMI value at lower layers, which means their patches have less query-adaptive attentions.

The reconstruction task requires holistic reasoning among patches to predict the masked signals and thus obtains semantic understanding to the input. Since this challenging task facilitates non-trivial inter-patch interactions, we apply it at multiple local layers, including both lower and upper ones to explicitly guide them all.

\subsection{Local multi-scale reconstruction}
In MIM, the supervision signals for a reconstruction task are directly calculated from the input. Concretely, we evenly divide the input image $x\in\mathbb{R}^{H\times W\times C}$ into non-overlapping regions $\{x_i\in\mathbb{R}^{p\times p\times C}\}_{i=1}^{HW/p^2}$ and use some feature descriptor $\pi$ to extract the supervision signal $y_i=\pi(x_i)$. To learn generalizing representations, many feature descriptors have been designed, such as pixel normalization \cite{he2022masked,xie2022simmim}, HOG \cite{wei2022masked} and the pre-trained or momentum teacher \cite{bao2021beit,dong2021peco,baevski2022data2vec,zhou2021image}. We define the scale of supervision $y$ as its spatial size $\frac{H}{p}\times\frac{W}{p}$. For a given input, the fine-scale supervisions from finely-divided input regions typically contain the low-level semantical information of the input, like corners, edges or textures. Relatively, the coarse-scale supervisions capture high-level semantical information of the input, like the shape of partial or whole object. Intuitively, multi-scale supervisions can guide representation learning better than the common single-scale ones due to their richer semantic information. In this work, we mainly consider the feature descriptors which are readily available without extra pre-training burden and costly forward inference of teacher networks, e.g., pixel normalization and HOG.

As shown in Table \ref{scale_vit}, directly applying the top-layer reconstruction task at carefully chosen local layers of ViT can not achieve meaningful improvement, where each local task predicts the supervisions of same scale. In fact, the lower and upper layers expect to learning low-level and high-level information respectively \cite{park2021vision,gao2022convmae}, so it is not appropriate to use the single-scale supervisions to guide multiple local layers, even for the columnar architectures with feature maps of the same scale at all layers. To this end, we make the lower layers to reconstruct fine-scale supervisions and the upper ones to reconstruct coarse-scale supervisions. Specially, for the pyramidal architectures which already hard-code the multi-scale property to the features by setting their spatial sizes, we use the supervisions with same scale as the feature maps at the chosen layers for compatibility.

The decoding process under a specific scale is illustrated in Fig. \ref{model} (b). The decoder consists of three parts: Transformer blocks for reasoning, (optional) Deconvolution/Pool for rescaling and Multilayer perceptron for prediction. Concretely, based on the encoded visible patches $z_v^l$ from $l$-th layer and the mask tokens $\{e_{[\mathbb{M}]}^l|m_i^l=1\}$, a decoder outputs prediction $\hat{y}^l$ that has the same scale as feature map $z_v^l$. When the supervision $y^l$ has different scale with the feature map $z_v^l$ (e.g., in the columnar architectures), the decoded prediction can not match the supervision. To this end, we use the deconvolution/pool operations to rescale the prediction $\hat{y}^l$ for matching the supervision $y^l$. For example, we can rescale the $14\times 14$ prediction to the scale of $56\times 56$ with twice deconvolution operations or to the scale of $7\times7$ with average pool. To avoid excessive computational overhead, we use tiny decoders containing one Transformer block with small embedding dimension.

The training loss is the weighted summation of the reconstruction losses at the chosen layers
\begin{equation}
	\mathcal{L}_{LocalMIM}=-\sum_{l\in \mathcal{I}}w_l\cdot\sum_{i=1}^{N_l}m_i^l\cdot \ln P(y_i^l|\hat{y}_i^l)
\end{equation}
where $\mathcal{I}$ is the set of chosen layers, $w_l$ is the coefficient of each local loss, and mask $m^l$ is calculated by up/down-sampling the initial mask $m$. These local losses guide the patches at multiple chosen layers to conduct semantic interactions under different scales, which not only accelerates the learning of multiple layers but also facilitates multi-scale semantical understanding to the input. As shown in Fig. \ref{nmi_vit}, compared with existing models, our LocalMIM has larger NMI values at lower layers, which means the attention maps depend more strongly on the query patches.

\section{Experiments}
In this section, we first evaluate our LocalMIM model on classification, detection and segmentation tasks, and then provide some ablation studies for deep understanding.

\begin{table}
	\centering
	\begin{tabular}{lccc}
		\toprule
		\textbf{Model} & \textbf{PT Epoch} & \textbf{PT Hours} & \textbf{mIoU} \\ 
		\midrule
		Supervised & - & - & 47.4 \\
		\hline
		MoCo v3 \cite{chen2021empirical} & 300 & -    & 47.3  \\
		BEiT \cite{bao2021beit} & 800 & 1920 & 47.1  \\
		MAE \cite{he2022masked} & 1600 & 1760 & 48.1 \\
		MaskFeat \cite{wei2022masked} & 1600 & 6240 & 48.8 \\
		PeCo \cite{dong2021peco} & 800 & - & 48.5 \\
		CAE \cite{chen2022context} & 800 & 2240 & 48.8 \\
		\rowcolor{green!15} LocalMIM-HOG & 1600 & 1120 & 49.5 \\
		\bottomrule
	\end{tabular}
	\caption{Semantic segmentation on ADE20K using UperNet with ViT-B backbone. Our LocalMIM achieves better results than previous MIM models with less pre-training duration.}
	\label{ADE20K}
	\vspace{-3mm}
\end{table}

\subsection{Classification on ImageNet-1K}
\paragraph{Settings.}
Both pre-training and fine-tuning are conducted on ImageNet-1K \cite{russakovsky2015imagenet} dataset under the $224\times224$ resolution. We mainly examine two representative architectures, columnar ViT \cite{DBLP:conf/iclr/DosovitskiyB0WZ21} and pyramidal Swin \cite{liu2021swin}. The input images are patchified with patch size $p=16$ for ViT and $p=4$ for Swin, and the obtained patches are randomly masked with ratio $r=0.75$ by default. We use both HOG feature \cite{wei2022masked} and normalized pixels \cite{he2022masked} as the supervision signals. For Swin, we introduce the reconstruction task after each stage and the supervision signals have the same scale with the output feature maps, e.g., the chosen layers $\mathcal{I}=\{2,4,22,24\}$ and the scales of supervision are $\{28^2,14^2,7^2,7^2\}$ on Swin-B. Each decoder contains one transformer block with the embedding dimension of 128 and 4 attention heads. For ViT, the chosen layers $\mathcal{I}=\{2,4,4+n,6+n\}$, e.g., $n=6$ on ViT-B and $n=18$ on ViT-L. The scales of supervision are $\{56^2,28^2,14^2,7^2\}$. Each decoder contains one transformer block with the embedding dimension of 256 and 8 attention heads. The pre-training and fine-tuning schedules mostly follow \cite{he2022masked,huang2022green}, and more detailed settings can be found in Appendix \ref{imple}.

\paragraph{Results.}
We compare our LocalMIM with existing MIM models and examine both pre-training efficiency and top-1 fine-tuning accuracy on ImageNet-1K. The results are provided in Fig. \ref{trade} and Table \ref{Finetune}. For fair comparison, we estimate the pre-training duration of each model on the same machine with one Tesla V100-32G GPU, CUDA 10.2 and Pytorch 1.8. We report the running time on single GPU, denoted as `GPU Hours', see Appendix \ref{GPU} for more details. On ViT-B, LocalMIM achieves the best results of MAE \cite{he2022masked} and MaskFeat \cite{wei2022masked} with $3.1\times$ and $5.6\times$ acceleration respectively. On Swin-B, LocalMIM achieves those of SimMIM$_{192}$ \cite{xie2022simmim} and GreenMIM \cite{huang2022green} with $3.6\times$ and $6.4\times$ acceleration respectively. In terms of final top-1 fine-tuning accuracy, LocalMIM achieves $84.0\%$ with ViT-B and $84.1\%$ with Swin-B. Compared with previous best results, LocalMIM achieves comparable performance with significantly less pre-training duration. Note that PeCo \cite{dong2021peco} uses more advanced feature descriptor, a pre-trained codebook, which introduces additional pre-training burden. Besides, on Swin-B, with 100 epochs pre-training and 100 epochs supervised fine-tuning LocalMIM achieves $83.8\%$ top-1 accuracy and takes about 350 GPU Hours, while 300 epochs supervised training from scratch only achieves $83.5\%$ top-1 accuracy and takes about 720 GPU Hours. This means even for the high-level classification task, self-supervision learning is more efficient and effective than supervised learning. Similar results can also be found on ViT.

\begin{table}
	\scalebox{0.9}{
		\centering
		\begin{tabular}{lcccc}
			\toprule
			\textbf{Model} & \textbf{PT Epoch} & \textbf{PT Hours} & \textbf{AP$^b$} & \textbf{AP$^m$} \\ 
			\midrule
			Supervised & 300 & 840 & 48.5 & 43.2 \\
			\hline
			SimMIM$_{192}$ \cite{xie2022simmim} & 800  & 1440 & 50.4 & 44.4 \\
			GreenMIM \cite{huang2022green}   & 800  & 640  & 50.0 & 44.1 \\
			\rowcolor{green!15} LocalMIM-HOG   & 400   & 440  & 50.7 & 44.9 \\
			\bottomrule
	\end{tabular}}
	\caption{Object detection and instance segmentation on COCO. We fine-tune Mask R-CNN end-to-end with Swin-B backbone.}
	\label{COCO}
	\vspace{-3mm}
\end{table}

\subsection{Downstream tasks}
We transfer our pre-trained backbone to semantic segmentation on ADE20K \cite{zhou2017scene} and object detection and segmentation on COCO \cite{lin2014microsoft}.

\begin{table*}[t]
	\centering
	\subfloat[
	\textbf{Reconstruction target}. HOG is more suitable as multi-scale supervisions than pixels.
	\label{target}]{
		\centering
		\begin{minipage}{0.29\linewidth}{\begin{center}
					\tablestyle{4pt}{1.05}
					\begin{tabular}{x{25}x{40}x{40}}
						target & ViT-B & Swin-B \\
						\shline
						baselines & 82.9  & 83.2 \\
						\hline
						Pixels  & 83.0  & 83.7 \\
						HOG    & \cellcolor{green!15} \textbf{83.3} & \cellcolor{green!15} \textbf{83.8} \\
					\end{tabular}
					\vspace{1.2em}
		\end{center}}\end{minipage}
	}
	\hspace{2em}
	\subfloat[
	\textbf{Mask ratio}. Masking $75\%$ patches works the best for both ViT and Swin.
	\label{ratio}]{
		\begin{minipage}{0.29\linewidth}{\begin{center}
					\tablestyle{4pt}{1.05}
					\begin{tabular}{x{25}x{40}x{40}}
						ratio & ViT-B & Swin-B \\
						\shline
						0.40 & 83.2 & 83.7 \\
						0.60 & 83.1 & 83.7 \\
						0.75 & \cellcolor{green!15} \textbf{83.3} & \cellcolor{green!15} \textbf{83.8} \\
						0.90 & 83.0 & 83.5 \\
					\end{tabular}
		\end{center}}\end{minipage}
	}
	\hspace{2em}
	\subfloat[
	\textbf{Decoder design}. A tiny decoder performs as well as the larger one but is more efficient.
	\label{decoder_design}]{
		\begin{minipage}{0.29\linewidth}{\begin{center}
					\tablestyle{1pt}{1.05}
					\begin{tabular}{y{50}x{40}x{40}}
						decoder    & ViT-B       & Swin-B      \\
						\shline
						512D - 16H & 83.3 (1.0h) & 83.8 (1.6h) \\
						256D - 8H  & \cellcolor{green!15} \textbf{83.3 (0.7h)} & 83.7 (1.3h)  \\
						128D - 4H  & 83.2 (0.6h)  & \cellcolor{green!15} \textbf{83.8 (1.1h)} \\
					\end{tabular}
				    \vspace{1.2em}
		\end{center}}\end{minipage}
	}
	
	\subfloat[
	\textbf{Locations for Swin}. Applying reconstructions at all stages mostly accelerates learning.
	\label{loc_swin}]{
		\begin{minipage}{0.29\linewidth}{\begin{center}
					\tablestyle{6pt}{1.05}
					\begin{tabular}{y{40}x{25}x{40}}
						locations     & backbone & acc \\
						\shline
						GreenMIM      & Swin-B & 83.2 \\
						\hline
						$[24]$        & \multirow{5}{*}{Swin-B} & 83.3 \\
						$[22,24]$     & & 83.4 \\
						$[4,22,24]$   & & 83.6 \\
						$[2,4,22,24]$ & & \cellcolor{green!15} \textbf{83.8} \\
						fusion        & & 83.5 \\
					\end{tabular}
		\end{center}}\end{minipage}
	}
	\hspace{2em}
	\subfloat[
	\textbf{Locations for ViT}. The chosen layers for local losses obviously affect performance.
	\label{pos_vit}]{
		\centering
		\begin{minipage}{0.29\linewidth}{\begin{center}
					\tablestyle{4pt}{1.05}
					\begin{tabular}{y{45}x{28}x{40}}
						locations & backbone & acc \\
						\shline
						MAE  & ViT-B & 82.9 \\
						\hline
						$[12]$ & \multirow{4}{*}{ViT-B} & 83.0 \\
						$[3,6,9,12]$  &  & 83.1 \\
						$[2,4,10,12]$ &  & \cellcolor{green!15} \textbf{83.3} \\
						$[1,2,11,12]$ &  & 82.9 \\
					\end{tabular}
				    \vspace{1.2em}
		\end{center}}\end{minipage}
	}
	\hspace{2em}
	\subfloat[
	\textbf{Scales for ViT}. Using the supervisions from fine-scale to coarse-scale works the best.
	\label{scale_vit}]{
		\begin{minipage}{0.29\linewidth}{\begin{center}
					\tablestyle{1pt}{1.05}
					\begin{tabular}{y{70}x{28}x{40}}
						scales & backbone & acc \\
						\shline
						MAE  & ViT-B & 82.9 \\
						\hline
						$[14^2,14^2,14^2,14^2]$ & \multirow{4}{*}{ViT-B} & 83.0 \\
						$[28^2,14^2,7^2,7^2]$   &  & 83.2 \\
						$[56^2,28^2,14^2,7^2]$  &  & \cellcolor{green!15} \textbf{83.3}  \\
						$[7^2,14^2,28^2,56^2]$  &  & 83.1 \\
					\end{tabular}
				    \vspace{1.1em}
		\end{center}}\end{minipage}
	}
	\caption{Ablation studies with ViT-B and Swin-B on ImageNet-1K. We report the top-1 fine-tuning accuracy ($\%$) and the \colorbox{green!15}{default} setting is: the reconstruction target is HOG feature, the mask ratio is $75\%$, the decoder contains one Transformer block with dimension 256 and 8 heads for ViT and dimension 128 and 4 heads for Swin, the local reconstructions are applied at all stages of Swin and $[2,4,10,12]$-th layer of ViT, the scales of supervision are $[28^2,14^2,7^2,7^2]$ for Swin and $[56^2,28^2,14^2,7^2]$ for ViT, and the pre-training length is 100 epochs.}
	\label{ablation}
	\vspace{-.5em}
\end{table*}

\paragraph{Semantic segmentation on ADE20K.}
We conduct semantic segmentation on ADE20K using UperNet \cite{xiao2018unified} and following the code in \cite{bao2021beit,he2022masked}. See Appendix \ref{imple} for fine-tuning details. The results are shown in Table \ref{ADE20K}. With significantly less computation burden, LocalMIM outperforms the state-of-the-art result by $0.7$. Besides, LocalMIM has the same top-1 fine-tuning accuracy with MaskFeat \cite{wei2022masked} on ImageNet-1K, but has better semantic segmentation performance, which means our local multi-scale reconstruction facilitates the learning of multi-scale semantic knowledge.

\paragraph{Object detection and instance segmentation on COCO.}
We fine-tune Mask R-CNN \cite{he2017mask} on COCO with Swin-B backbone. Following \cite{huang2022green}, we also use the code base and $3\times$ fine-tuning schedule from the supervised Swin, where the model is fine-tuned for 36 epochs. See Appendix \ref{imple} for fine-tuning details. We report box AP (AP$^b$) for object detection and mask AP (AP$^m$) for instance segmentation. The results are shown in Table \ref{COCO}. With no labeling burden, our LocalMIM outperforms supervised pre-training by 2.2 AP$^b$ and 1.7 AP$^m$. Compared to SimMIM$_{192}$ and GreenMIM, our LocalMIM achieves $+$0.3 and $+$0.7 gain respectively for detection, and $+$0.5 and $+$0.8 gain for segmentation with significantly less pre-training duration.

\subsection{Ablation studies}
We ablate our LocalMIM using the default setting in Table \ref{ablation}. MAE \cite{he2022masked} and GreenMIM \cite{huang2022green} are our main baselines on ViT and Swin respectively, since we all adopt the asymmetric encoder-decoder strategy.

\paragraph{Reconstruction target.}
We examine two types of signal as supervisions: HOG feature \cite{wei2022masked} and normalized pixel \cite{he2022masked}, and the results are shown in Table \ref{target}. HOG feature is more suitable as multi-scale supervisions than normalized pixel on both ViT and Swin, since it contains more refined semantic information. Compared to the columnar ViT with single-scale feature maps, our multi-scale supervisions achieve more improvement on the pyramidal Swin. Since we need to rescale the predictions in the columnar architectures, it is harder to guide the learning of multi-scale information, e.g., we need more information-refined signals (HOG) and obtain relatively less improvement.

\paragraph{Mask ratio.}
We explore different mask ratios, ranging from 0.4 to 0.9, and the results are provided in Table \ref{ratio}. Similar to \cite{he2022masked,huang2022green}, our LocalMIM is robust to the mask ratio from 0.4 to 0.75, and excessively large one (0.9) harms the performance since it over-complicates the reconstruction tasks.

\paragraph{Decoder design.}
As shown in \cite{he2022masked,huang2022green}, a small single-block decoder can achieve the similar fine-tuning accuracy to the heavier one with multiple blocks. However, since we need to use the decoder for each reconstruction task to process the signals of specified scale, their decoder which has the embedding dimension of 512 and 16 attention heads is still heavy. To this end, we explore whether a smaller decoder can still handle the local reconstruction tasks for learning good representations. One Transformer block is the minimal requirement to propagate information from visible patches to mask tokens, so we still use one Transformer block but narrow the embedding dimension and reduce the number of attention heads. The results (top-1 fine-tuning accuracy and GPU Hours per epoch) are shown in Table \ref{decoder_design}, where `$d$D - $h$H' represents the decoder with the embedding dimension of $d$ and $h$ attention heads. Surprisingly, a tiny decoder can achieve the same even better fine-tuning performance, but is significantly more efficient.

\paragraph{Locations for local reconstructions.}
An important design of our LocalMIM is the locations in the encoder where we introduce local reconstructions. Many pyramidal architectures like Swin already divide the entire backbone into multiple stages, so we introduce multi-scale reconstructions at their output locations. From the last stage, we gradually include the lower stages and the results are shown in Table \ref{loc_swin}. Note that the $2$-th, $4$-th, $22$-th and $24$-th layer correspond to the output location of the four stages respectively. We also provide the result of fusion method \cite{gao2022convmae} which fuses the output features from each stage for reconstruction. As we can see, introducing reconstructions at local stages can accelerate representation learning, and applying to all stages works the best. Our LocalMIM also outperforms the fusion method which only uses the single-scale reconstruction, and this verifies the importance of multi-scale reconstructions. Further, for the columnar ViT, we follow the experience from Swin and also select four local layers. Concretely, we consider the uniform division $[3,6,9,12]$, Swin-style division $[2,4,10,12]$ and a extreme division $[1,2,11,12]$. The results are shown in Table \ref{pos_vit}. The performance on columnar ViT is sensitive to the chosen local layers, and the Swin-style division $[2,4,4+n,6+n]$ works the best.

\paragraph{Scales of supervisions.}
Since the pyramidal architectures hard-code the multi-scale property of the feature maps, we directly use the supervisions of same scale as the feature maps for compatibility, i.e., we use $[28^2,14^2,7^2,7^2]$ for Swin. For columnar ViT, we consider single-scale supervisions $[14^2,14^2,14^2,14^2]$, fine-to-coarse multi-scale supervisions $[28^2,14^2,7^2,7^2]$ and $[56^2,28^2,14^2,7^2]$, and coarse-to-fine multi-scale supervisions $[7^2,14^2,28^2,56^2]$. The results are shown in Table \ref{scale_vit}. The multi-scale supervisions achieve better fine-tuning accuracy than the single-scale ones, which means the multiple local layers require the supervision of different scales. Further, the fine-to-coarse supervisions outperform the coarse-to-fine ones, which is consistent with the design priors for visual architectures \cite{liu2021swin} and biological visual processing \cite{hubel1960receptive}.

\begin{figure}
	\begin{center}
		\includegraphics[width=1.0\linewidth]{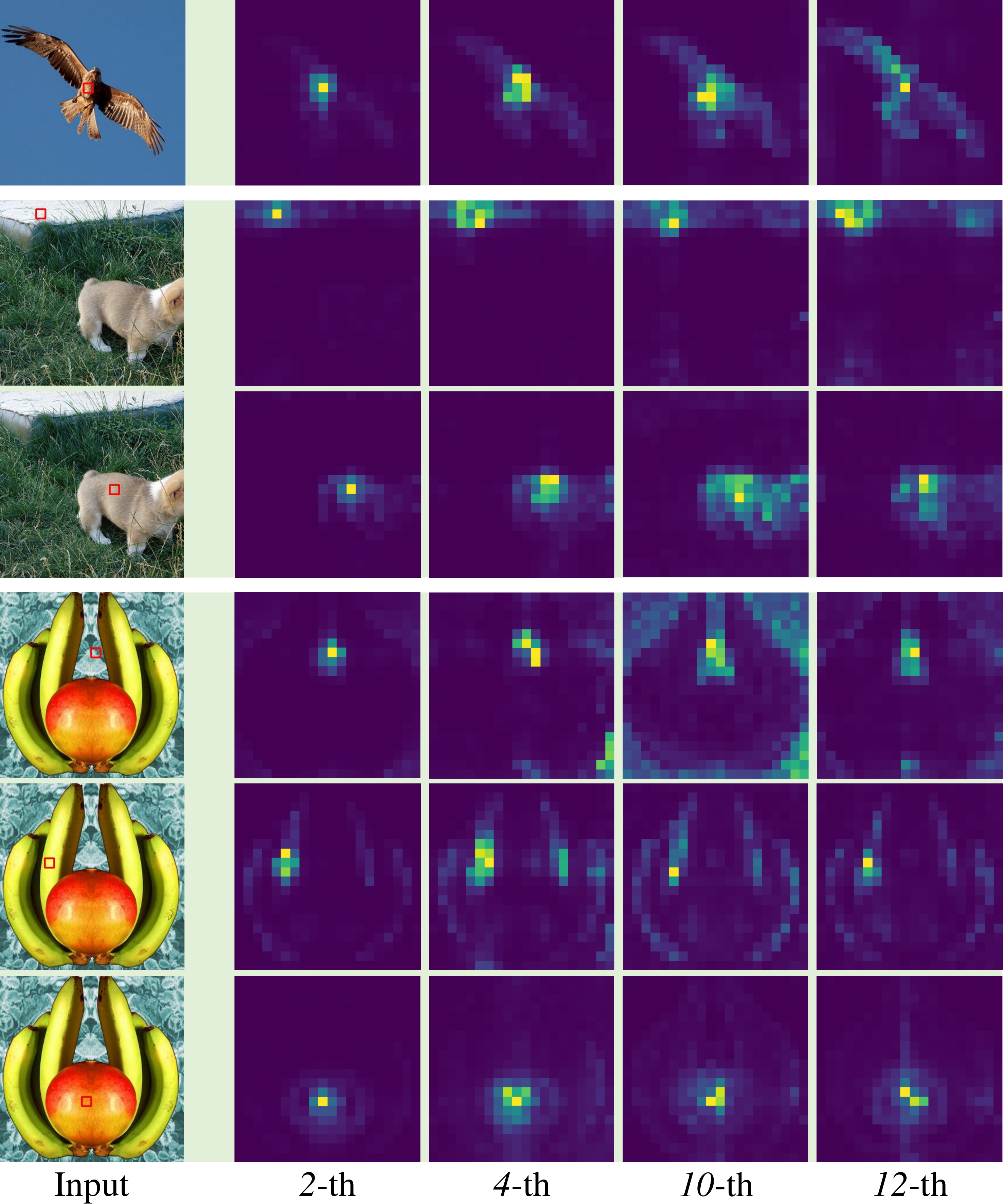}
	\end{center}
	\vspace{-3mm}
	\caption{Visualization of the attention maps for different query points, marked with red boxes. LocalMIM can distinguish semantical regions using self-attention mechanism.}
	\label{attention}
	\vspace{-4mm}
\end{figure}

\paragraph{Analysis of self-attention mechanism.}
To qualitatively illustrate the behavior of self-attention mechanism in LocalMIM, we visualize the attention maps for different query positions within an image, as shown in Fig. \ref{attention}. We use each chosen patch as query and show the patches it attends to. We select the attention maps at 2-th, 4-th, 10-th and 12-th layers of a pre-trained ViT-B/16 backbone to show their changes during forward inference. For object-centric images, LocalMIM can distinguish the foreground object from the background. For more complex multi-object images, LocalMIM can effectively separate different objects without any task-specific supervision, which means the attention maps are query-adaptive. On the other hand, the patches at lower layers typically more focus on their neighboring regions and capture low-level information, while those at upper layers attend to a wide range of semantically related regions and capture high-level shape information.

\paragraph{Gradient-isolated training.}
Inspired by locally supervised learning \cite{lowe2019putting,nokland2019training}, we remove global back-propagation and stop the gradients after each chosen local layer used for reconstruction. The entire backbone is thus divided into multiple gradient-isolated parts. The results are shown in Table \ref{greedy}. Surprisingly, the gradient-isolated training achieves similar performance to global back-propagation, which further verifies the effectiveness of our multi-scale reconstructions for guiding the local layers. It even requires no gradient information from the upper layers. This observation also shows the promise of our local multi-scale reconstruction for the decoupled training of neural networks, which allows for training very deep networks without memory concern and reduces explosive or vanishing gradients.

\begin{table}
	\centering
		\begin{tabular}{lccc}
			model & backbone & GPU Hours/Ep. & acc \\ 
			\shline
			LocalMIM    & \multirow{2}{*}{ViT-B} & 0.7 & 83.3  \\
			w/ isolated grad &  & 0.7  & 83.0  \\
			\hline
			LocalMIM    & \multirow{2}{*}{Swin-B} & 1.1  & 83.8 \\
			w/ isolated grad &  & 1.1 & 83.7 \\
	\end{tabular}
	\caption{Training LocalMIM with isolated gradients achieves similar performance with global back-propagation.}
	\label{greedy}
	\vspace{-3mm}
\end{table}

\section{Conclusions}
We present a novel and efficient pretext task, \emph{local multi-scale reconstruction}, where the lower layers and upper layers reconstruct the fine-scale and coarse-scale supervisions from the input respectively. We obtain the multi-scale supervisions by first dividing the input under different scales and then extracting supervisions with appropriate feature descriptors. Our model needs no extra pre-trained codebook and no costly forward inference of teacher networks during pre-training. The asymmetric encoder-decoder strategy with tiny encoders also have small computational burden, our model thus can be trained quickly. The novel pretext task further accelerates the representation learning, especially for the pyramidal architectures. All these designs allow our model to achieve comparable performance to existing models but with significantly less pre-training burden. The gradient-isolated training also verifies the effectiveness of our novel task design in guiding the local layers.

\paragraph{Acknowledgment.} We gratefully acknowledge the support of MindSpore \cite{mindspore}, CANN (Compute Architecture for Neural Networks) and Ascend AI Processor.

{\small
\bibliographystyle{ieee_fullname}
\bibliography{egbib}
}

\clearpage
\appendix

\section{More experiments}
In this section, we provide more experiments to support our work.

\subsection{Decoupling local and multi-scale}
To effectively guide the local layers, we propose multi-scale supervisions for multiple local reconstruction tasks. Here we decouple the local (or global) reconstruction and multi-scale (or single-scale) supervisions to further understand their relations. For global multi-scale reconstruction, we conduct at the top layer of encoder and use separate decoders to predict multiple supervisions of different scales. When the predictions have different scale with supervisions, we use deconvolution/pooling options to rescale them for matching supervisions. The results are shown in Table \ref{dcp} and the pre-training length is 100 epochs. As we can see, global reconstruction prefers to single-scale supervisions, and using the supervisions of different scales to guide the same layer could make confusion. Conversely, local reconstruction prefers to multi-scale supervisions, and multiple local layers expect to learn the information of different scales. Local reconstruction can achieve better performance than the global one in most cases, and the gain increases when using multi-scale supervisions.

\begin{table}
	\centering
	\subfloat[
	ViT-B
	\label{dcp_vit}]{
		\centering
		\tablestyle{1pt}{1.2}
		\begin{tabular}{x{25}|x{40}x{40}}
			& single-scale & multi-scale  \\
			\shline
			global & 83.0         & 82.8   \\
			local  & 83.0         & 83.3 \\
	\end{tabular}}
	\subfloat[
	Swin-B
	\label{dcp_swin}]{
		\centering
		\tablestyle{1pt}{1.2}
		\begin{tabular}{x{25}|x{40}x{40}}
			& single-scale & multi-scale  \\
			\shline
			global & 83.3         & 82.9   \\
			local  & 83.6       & 83.8  \\
	\end{tabular}}
	\caption{Decoupling local reconstruction and multi-scale supervisions. We report the top-1 fine-tuning accuracy on ImageNet-1K.}
	\label{dcp}
\end{table}

\subsection{Comparison with feature fusion}
\begin{figure*}
	\centering
	\subcaptionbox{ViT-B}{\includegraphics[width=0.45\textwidth]{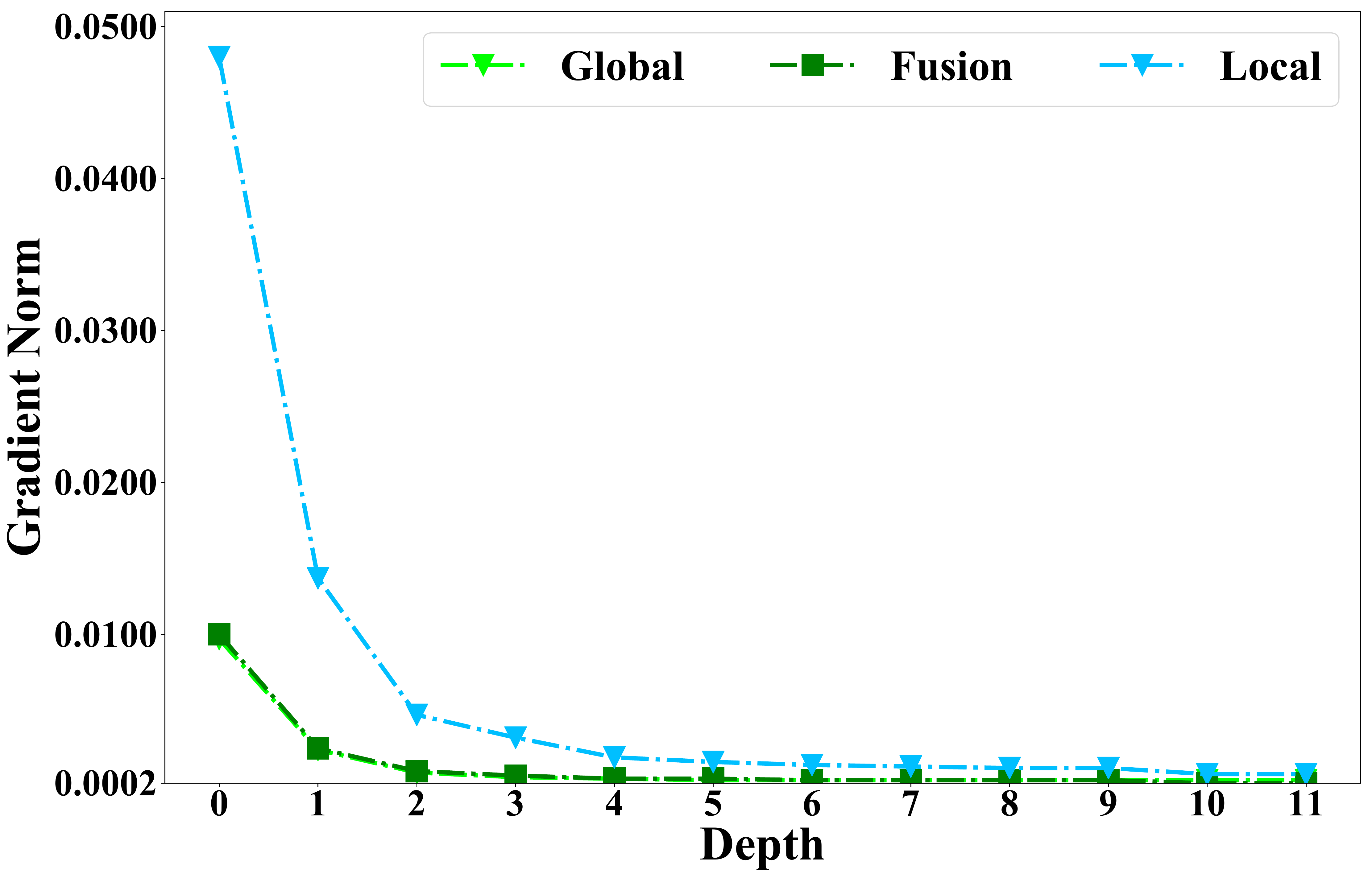}}
	\subcaptionbox{Swin-B}{\includegraphics[width=0.45\textwidth]{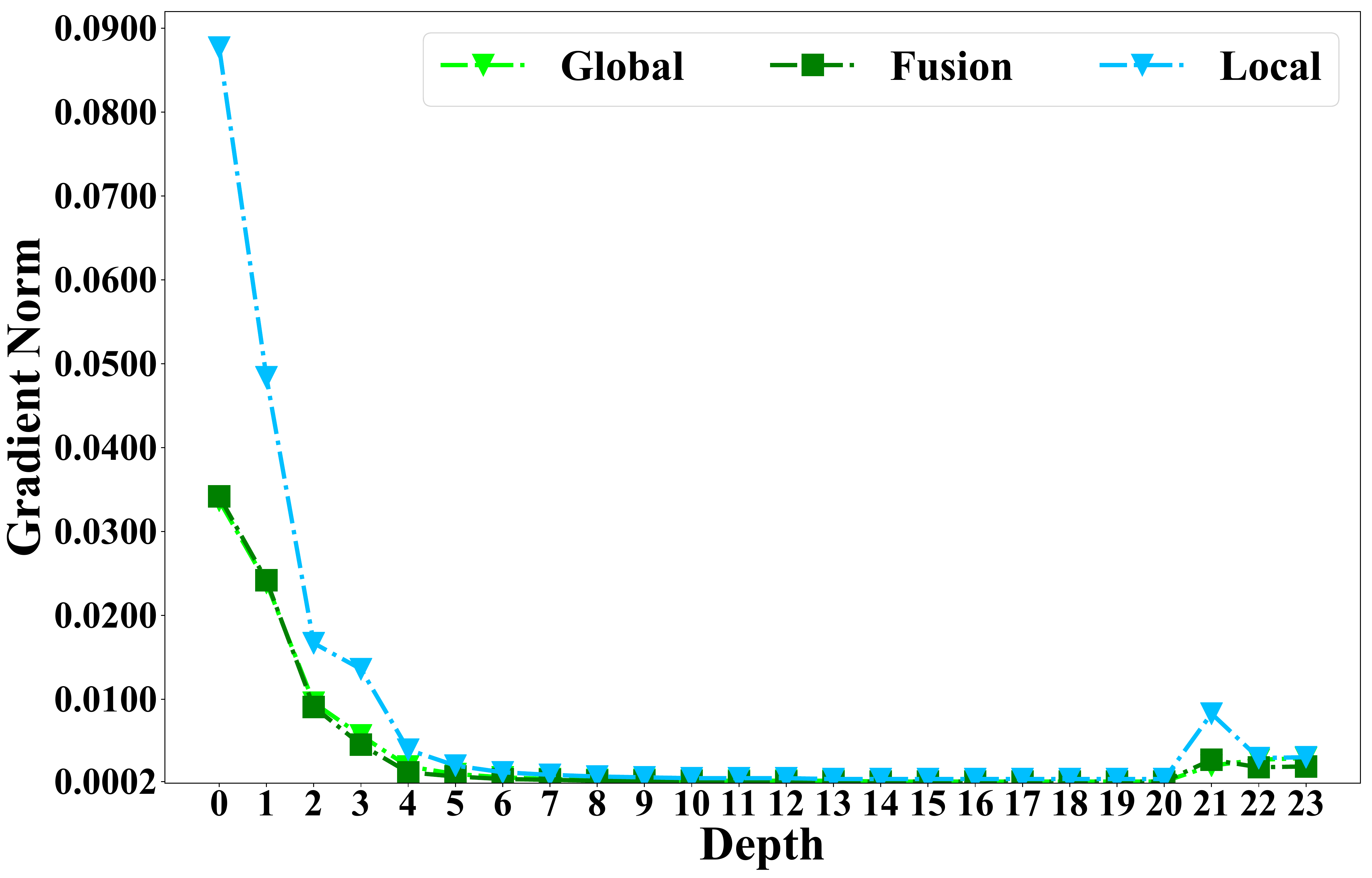}}
	\caption{Gradient norm of each layer in the encoder. We compare the global reconstruction, the global reconstruction with feature fusion and our local reconstruction, which are denoted as `Global', `Fusion' and `Local' respectively.}
	\label{grad}
\end{figure*}

To explicitly guide the lower layers, we conduct reconstruction task at multiple chosen local layers. The other method is fusing the features of multiple local layers to the top layer for global reconstruction \cite{gao2022convmae}. It uses single-scale supervision for avoiding confusion. We compare our local reconstruction with this feature fusion method, and the results are shown in Table \ref{fus}. Local reconstruction achieves consistently better performance than feature fusion on both columnar ViT \cite{DBLP:conf/iclr/DosovitskiyB0WZ21} and pyramidal Swin \cite{liu2021swin}. For further exploration, we examine the gradient norm of each layer in the encoder during training process. Concretely, we load the checkpoint (state) of the median epoch in a complete training schedule and then calculate the gradient norm of parameters in each layer under this state. The results are shown in Fig. \ref{grad}. For other middle epochs, we observe the same results. The lower layers have larger gradient norm than the upper ones due to the skip-connections in vision transformers. The skip-connections allow the lower layers to learn more quickly than the upper ones, which may be one reason for its significant effectiveness in various architectures \cite{he2016deep,vaswani2017attention,yu2022metaformer}. Our local reconstruction can strengthen this characteristic and thus obtain better performance. Feature fusion essentially has the similar effect with the skip-connections. Besides, another advantage of our local construction is that it is compatible with multi-scale supervisions and thus can take advantage of richer information.

\begin{table}
	\centering
	\subfloat[
	ViT-B
	\label{fus_vit}]{
		\centering
		\tablestyle{2pt}{1.2}
		\begin{tabular}{x{70}|x{25}}
			method                & acc   \\
			\shline
			global                & 83.0  \\
			global with fusion    & 83.0  \\
			local                 & 83.3  \\
	\end{tabular}}
	\hspace{2mm}
	\subfloat[
	Swin-B
	\label{fus_swin}]{
		\centering
		\tablestyle{2pt}{1.2}
		\begin{tabular}{x{70}|x{25}}
			method                & acc   \\
			\shline
			global                & 83.3  \\
			global with fusion    & 83.5  \\
			local                 & 83.8  \\
	\end{tabular}}
	\caption{Top-1 fine-tuning accuracy on ImageNet-1K. We compare the global reconstruction, the global reconstruction with feature fusion and our local reconstruction.}
	\label{fus}
\end{table}

\subsection{Query-adaptive attention}
\begin{figure}
	\centering
	\includegraphics[width=0.85\linewidth]{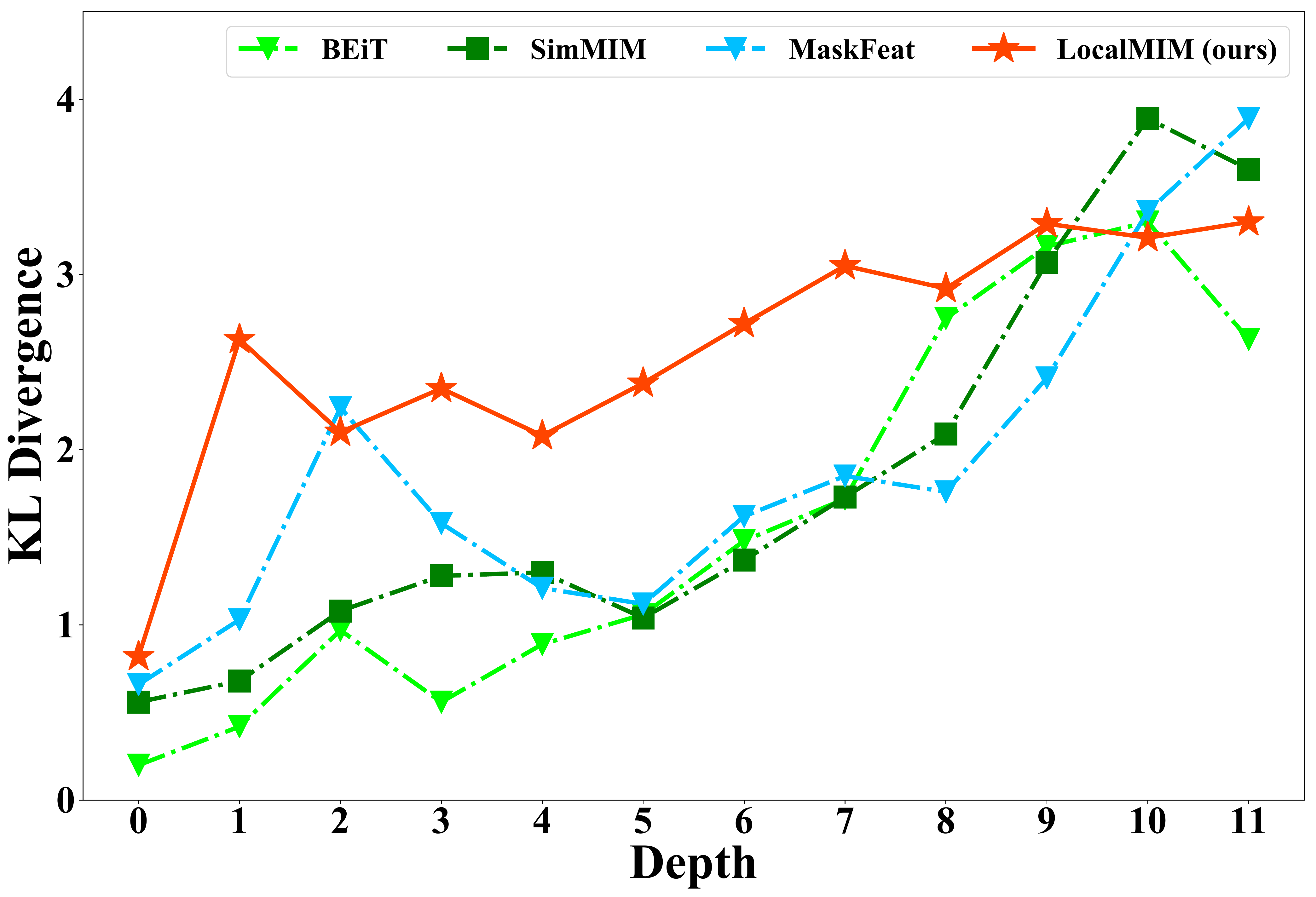}
	\caption{The KL divergence between attention distributions of different query patches at each layer of a pre-trained ViT-B backbone, averaged on all pairs of query patches.}
	\label{kl}
\end{figure}

In the main text, we use Normalized Mutual Information (NMI) \cite{strehl2002cluster} between query and key patches to examine how much the attention map depends on the query patch. Here we use another metric, the Kullback-Leibler divergence between the attention distributions of different query patches. Intuitively, when the attention map strongly depends on the query patch, the attention distributions of a pair of query patches should have large KL divergence. We calculate the average on all pairs of query patches at each layer and the results are shown in Fig. \ref{kl}. As we expect, existing MIM models with global loss have small KL divergence at lower layers, which means the patches there have less query-adaptive attention. Relatively, the lower layers in our LocalMIM have larger KL divergence and the attention maps depend more strongly on the query patches.

\section{GPU Hours}\label{GPU}
`GPU Hours' denotes the running time on single Tesla V100-32G GPU. For fair comparison, we estimate that of each model at the same machine with one Tesla V100-32G GPU, CUDA 10.2 and PyTorch 1.8. We pre-train each model for $10$ epochs using its official released codes and default hyper-parameters, and then calculate the average running time per epoch. We find that each epoch takes similar time with each other during estimation, so pre-training $10$ epochs is enough to estimate the GPU Hours per epoch. The batch size is an important factor that affects the running time, and we choose it from $\{32,48,64,128,256\}$ to take full advantage of GPU memory and computing capability. This estimation method avoids the interference of the communication time among multiple GPUs.

\section{Implementation details}\label{imple}
For ViT \cite{DBLP:conf/iclr/DosovitskiyB0WZ21}, we use the standard architecture with the sine-cosine positional embeddings and do not use relative positional encoding or layer scaling. For HOG feature, we set the number of orientation bins $\#bins=18$ and the cell size is the same as the divided regions. We set the same weight to each local loss for simplicity. The pre-training and fine-tuning schedules mostly follow \cite{he2022masked,huang2022green}.

\begin{table}
	\centering
	\scalebox{0.9}{
		\begin{tabular}{l|cc}
			config & ViT & Swin  \\ 
			\shline
			optimizer              & \multicolumn{2}{c}{AdamW \cite{DBLP:conf/iclr/LoshchilovH19}} \\
			base learning rate     & $2e^{-4}$ & $1e^{-4}$ \\
			weight decay           & \multicolumn{2}{c}{0.05}   \\
			optimizer momentum     & \multicolumn{2}{c}{$\beta_1,\beta_2=0.9,0.95$}  \\
			batch size             & \multicolumn{2}{c}{2048 (B) / 4096 (L)}  \\
			learning rate schedule & \multicolumn{2}{c}{cosine decay \cite{DBLP:conf/iclr/LoshchilovH17}}  \\
			augmentation           & \multicolumn{2}{c}{RandomResizedCrop}  \\
			input resolution       & \multicolumn{2}{c}{$224\times224$}  \\
	\end{tabular}}
	\caption{Pre-training setting on ImageNet-1K.}
	\label{pre-train}
\end{table}

\begin{table}
	\centering
	\scalebox{0.8}{
		\begin{tabular}{l|cccc}
			\multirow{2}{*}{config} & \multicolumn{2}{c}{ViT} & \multicolumn{2}{c}{Swin}  \\ 
			& ViT-B & ViT-L & Swin-B & Swin-L  \\ 
			\shline
			optimizer              & \multicolumn{4}{c}{AdamW} \\
			peak learning rate     & \multicolumn{2}{c}{$\{2e^{-3},3e^{-3},4e^{-3}\}$} & \multicolumn{2}{c}{$\{3e^{-3},4e^{-3},5e^{-3}\}$} \\
			weight decay           & \multicolumn{4}{c}{0.05}   \\
			optimizer momentum     & \multicolumn{4}{c}{$\beta_1,\beta_2=0.9,0.999$}  \\
			layer-wise lr decay \cite{DBLP:conf/iclr/ClarkLLM20} & \multicolumn{2}{c}{$\{0.65,0.75\}$} & \multicolumn{2}{c}{$\{0.80,0.90\}$}  \\
			batch size             & \multicolumn{4}{c}{1024 (B) / 4096 (L)}  \\
			learning rate schedule & \multicolumn{4}{c}{cosine decay}  \\
			fine-tuning epochs     & 100 & 50 & 100 & 100 \\
			warmup epochs          & 20 & 5 & 20 & 20  \\
			drop path \cite{huang2016deep} & 0.1 & 0.2 & 0.1 & 0.3  \\
			augmentation           & \multicolumn{4}{c}{RandAug (9, 0.5) \cite{cubuk2020randaugment}}  \\
			label smoothing \cite{szegedy2016rethinking} & \multicolumn{4}{c}{0.1}  \\
			mixup \cite{DBLP:conf/iclr/ZhangCDL18} & \multicolumn{4}{c}{0.8}  \\
			cutmix \cite{yun2019cutmix} & \multicolumn{4}{c}{1.0}  \\
			input resolution       & \multicolumn{4}{c}{$224\times224$}  \\
	\end{tabular}}
	\caption{Fine-tuning setting on ImageNet-1K.}
	\label{fine-tune}
\end{table}

\paragraph{Pre-training.}
The default setting is shown in Table \ref{pre-train}. We use the simple data augmentation and do not use drop path or gradient clip. We use the linear learning rate scaling rule \cite{goyal2017accurate}: $lr=base\_lr\times batch\_size/256$. The warmup epoch \cite{goyal2017accurate} is set to 10 for pre-training 100 epochs, 40 for pre-training 400, 800 and 1600 epochs.

\paragraph{Fine-tuning on ImageNet-1K.}
The default fine-tuning setting is shown in Table \ref{fine-tune}. Most of the hyper-parameters are shared, except the peak learning rate, layer-wise learning rate decay and drop path rate, which are influenced by the backbones and the number of pre-training epochs.

\paragraph{Semantic segmentation on ADE20K.}
We use UperNet \cite{xiao2018unified} with ViT-B backbone and follow the semantic segmentation code of \cite{bao2021beit,he2022masked}. Concretely, we fine-tune end-to-end for 160K iterations using AdamW optimizer with the peak learning rate of $4e^{-4}$, weight decay of 0.05 and batch size of 16. The learning rate warmups with 1500 iterations and then decays with linear strategy. The model is trained with input resolution of $512\times512$ and uses bilinear positional embedding interpolate. We choose the out indices of feature maps as $[2,4,10,12]$ and use FPN \cite{lin2017feature} to rescale them.

\paragraph{Object detection and segmentation on COCO.}
We fine-tune Mask R-CNN \cite{he2017mask} on COCO \cite{lin2014microsoft} with Swin-B backbone. Following \cite{huang2022green}, we also use the code base and schedule from \cite{liu2021swin}. Concretely, the model is fine-tuned on COCO 2017 train split and evaluated on 2017 val split. We adopt the $3\times$ fine-tuning schedule which trains the model for 36 epochs in total and decays the learning rate at the 27-th and 33-th epoch by a factor of 10. We use AdamW optimizer with the learning rate of $1e^{-4}$ and weight decay of $0.05$.

\end{document}